\documentclass[11pt]{article}
\usepackage{amsmath}
\usepackage{amssymb}
\usepackage{textcomp}
\usepackage{enumitem}
\usepackage{booktabs}

\usepackage{listings}
\lstdefinelanguage{json}{
    morestring=[b]",
    morestring=[s]{'}{'},
    morecomment=[l]{//},
    morecomment=[s]{/*}{*/},
    morekeywords={true,false,null},
    sensitive=false
}

\usepackage[final]{acl}

\usepackage{times}
\usepackage{latexsym}

\usepackage[T1]{fontenc}
\usepackage[utf8]{inputenc}
\usepackage{CJKutf8}
\usepackage{multirow}
\usepackage{microtype}
\usepackage{inconsolata}
\usepackage{graphicx}
\usepackage{caption}
\usepackage[many]{tcolorbox}

\usepackage{algorithm}
\usepackage{algorithmic}
\usepackage{float}

\makeatletter
\newwrite\app@toc
\newif\ifapp@tocwrite
\app@tocwritefalse
\newcommand{\app@writecontentsline}[3]{%
    \begingroup
        \protected@write\app@toc{}{%
            \string\contentsline{#2}{#3}{\thepage}{}%
        }%
    \endgroup
}

\AtBeginDocument{%
  \let\app@addcontentsline@orig\addcontentsline
  \renewcommand{\addcontentsline}[3]{%
      \ifapp@tocwrite
          \def\app@tmp{#1}%
          \def\app@tocname{toc}%
          \ifx\app@tmp\app@tocname
              \app@writecontentsline{#1}{#2}{#3}%
          \else
              \app@addcontentsline@orig{#1}{#2}{#3}%
          \fi
      \else
          \app@addcontentsline@orig{#1}{#2}{#3}%
      \fi
  }%
}

\newcommand{\app@starttoc}{%
    \IfFileExists{\jobname.apptoc}{%
        \begingroup
            \@ifpackageloaded{hyperref}{%
                \InputIfFileExists{\jobname.apptoc}{}{}%
            }{%
                \let\app@oldcontentsline\contentsline
                \def\contentsline####1####2####3####4{\app@oldcontentsline{####1}{####2}{####3}}%
                \InputIfFileExists{\jobname.apptoc}{}{}%
            }%
        \endgroup
    }{%
    }%
}
\AtEndDocument{\immediate\closeout\app@toc}
\makeatother

\usepackage[table]{xcolor}
\usepackage{framed}

\definecolor{codegreen}{rgb}{0,0.6,0}
\definecolor{codegray}{rgb}{0.5,0.5,0.5}
\definecolor{codepurple}{rgb}{0.58,0,0.82}
\definecolor{backcolour}{rgb}{0.95,0.95,0.92}
\definecolor{promptback}{rgb}{0.9,0.9,0.9}

\lstdefinestyle{jsonstyle}{
    backgroundcolor=\color{backcolour},
    commentstyle=\color{codegreen},
    keywordstyle=\color{magenta},
    numberstyle=\tiny\color{codegray},
    stringstyle=\color{codepurple},
    basicstyle=\ttfamily\footnotesize,
    breakatwhitespace=false,
    breaklines=true,
    captionpos=b,
    keepspaces=true,
    numbers=left,
    numbersep=5pt,
    showspaces=false,
    showstringspaces=false,
    showtabs=false,
    tabsize=2,
    language=json
}
\title{LatentRefusal: Latent-Signal Refusal for Unanswerable Text-to-SQL Queries}

\author{
  Xuancheng Ren \quad Shijing Hu \quad Zhihui Lu\thanks{\ \ Corresponding author.} \quad Jiangqi Huang \\
  Fudan University \\
  \texttt{\{xcren25, sjhu24, 25213050189\}@m.fudan.edu.cn, lzh@fudan.edu.cn} \\
  \AND
  Qiang Duan \\
  Pennsylvania State University Abington College \\
  \texttt{qduan@psu.edu}
}

\begin{document}
\maketitle
\begin{abstract}
In LLM-based Text-to-SQL systems, unanswerable and underspecified user queries may generate not only incorrect text but also executable programs that yield misleading results or violate safety constraints, thus posing a major barrier to safe deployment. Existing refusal strategies for such queries either rely on output-level instruction following, which is brittle due to model hallucinations, or on estimating output uncertainty, which adds complexity and overhead. To address this challenge, we first formalize safe refusal in Text-to-SQL systems as an answerability-gating problem, and then propose \textsc{LatentRefusal}, a latent-signal refusal mechanism that predicts query answerability from intermediate hidden activations of an LLM. We introduce the Tri-Residual Gated Encoder (TRGE), a lightweight probing architecture, to suppress schema noise and amplify sparse, localized question--schema mismatch cues that indicate unanswerability. Extensive empirical evaluations across diverse ambiguous and unanswerable settings, together with ablations and interpretability analyses, demonstrate the effectiveness of the proposed scheme and show that \textsc{LatentRefusal} provides an attachable, efficient safety layer for Text-to-SQL systems. Across four benchmarks, \textsc{LatentRefusal} achieves an average F1 of 88.5\% and 88.8\% on Llama-3.1-8B and Qwen-3-8B respectively, while adding $\sim$2ms probe overhead.
\end{abstract}

\section{Introduction}
Large language models (LLMs) have broadened access to data analytics by translating natural language questions into executable SQL (Text-to-SQL) \cite{sun2024sqlpalm,gao2024texttoSQL,li2025survey}. However, in real deployments (e.g., finance, healthcare, security), practical adoption is constrained by a safety-critical failure mode: unreliable behavior under \emph{unanswerable} or \emph{underspecified} queries. Such queries may require non-existent schema elements, admit multiple plausible interpretations, fall outside the database scope, or depend on subjective criteria. When optimized for helpfulness, LLMs can still produce plausible-looking SQL that is semantically incorrect. Unlike open-ended dialogue, where hallucinations mainly yield incorrect text, Text-to-SQL hallucinations yield \emph{executable programs}, which can silently corrupt reports, trigger privacy violations, or cause costly operational incidents \cite{huang2023survey,ji2023survey}.

\begin{figure}[t]
    \centering
    \includegraphics[width=\linewidth]{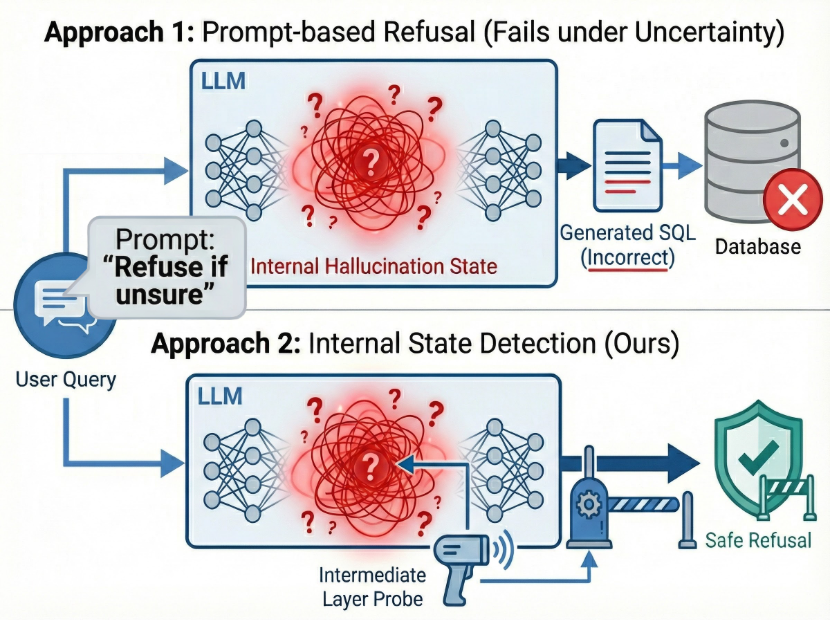}
    \caption{Comparison of refusal paradigms in Text-to-SQL systems. \textbf{Top:} Traditional prompt-based methods rely on the LLM's decoded output to decide refusal, which is brittle under uncertainty and can fail when the model hallucinates plausible but incorrect SQL. \textbf{Bottom:} \textsc{LatentRefusal} detects refusal signals directly from the frozen LLM's intermediate hidden states \emph{before} any SQL generation, enabling a single-pass, deterministic, and low-latency safety gate without generating or executing potentially harmful queries.}
    \label{fig:intro_comparison}
    \vspace{-1em}
\end{figure}

Throughout this paper, we define \emph{answerable} queries as those resolvable to a unique, valid SQL statement given the schema. Conversely, we use \emph{unanswerable} to broadly encompass queries suffering from missing schema elements, out-of-scope requests, subjectivity, or linguistic ambiguity. In these contexts, \emph{Text-to-SQL hallucination} refers to the generation of executable but semantically unfaithful SQL, a frequent risk when models are forced to answer unanswerable inputs.

Accordingly, a production Text-to-SQL system must be able to \emph{refuse} when it cannot answer safely and faithfully. Existing refusal strategies, however, face a persistent trade-off between reliability and efficiency. \emph{Prompt-based} methods instruct the model to abstain (e.g., ``refuse if unsure''), but refusal behavior is brittle under uncertainty and can fail precisely when the model hallucinates. In contrast, \emph{uncertainty-based} methods (e.g., self-consistency or semantic-entropy style scoring) can be more robust, yet often require multiple samples and substantial inference overhead. Worse, in Text-to-SQL, assessing agreement among candidate programs frequently depends on execution to resolve semantic equivalence---but executing questionable SQL to \emph{estimate} uncertainty undermines the goal of safety gating.

We observe that semantic information is implicit in the intermediate layers, which can be leveraged for refusal judgment. Meanwhile, experiments show that the middle layers contain the most accurate refusal direction information \cite{skean2025layer}.
Accordingly, we propose \textsc{LatentRefusal}, a refusal mechanism that makes the safety decision \emph{before generation} by detecting refusal signals directly from a frozen LLM's internal hidden states. As illustrated in Figure~\ref{fig:intro_comparison}, rather than relying on the model's final output (top), \textsc{LatentRefusal} attaches a light\-weight inter\-me\-diate-layer probe to predict answer\-ability in a single forward pass (bottom). This design yields a de\-ter\-min\-is\-tic, low-latency refusal gate that avoids sampling, avoids generating potentially harmful SQL, and avoids executing any SQL during the decision process.

{\sloppy
A key challenge is that naive probing over pooled hidden states is often insufficient for schema-con\-ditioned Text-to-SQL: inputs can be dominated by lengthy schema de\-scrip\-tions, while refusal cues (e.g., a missing column, absent join path, or under\-specified constraint) are subtle and localized. To address this, we introduce the Tri-Residual Gated Encoder (TRGE), a probe ar\-chi\-tec\-ture designed to suppress schema noise and amplify question--schema mismatch signals indicative of un\-answer\-ability.\par}

Our contributions are as follows:
\begin{itemize}[leftmargin=*]
    \setlength\itemsep{0em}
    \item \textbf{Problem and constraint formulation for safe refusal in Text-to-SQL.} We formalize refusal as an \emph{answerability gating} problem under a strict safety constraint: the system must decide whether to answer \emph{before} generating or executing any SQL, avoiding execution-based uncertainty estimation.
    \item \textbf{\textsc{LatentRefusal}: Single-pass latent-signal refusal.} We propose a mechanism that predicts answerability directly from a frozen LLM's hidden activations. This enables deterministic, low-latency refusal decisions in a single forward pass, avoiding the overhead of sampling or unsafe query execution.
    \item \textbf{TRGE: A probe for sparse refusal cues in schema-heavy prompts.} We introduce the Tri-Residual Gated Encoder (TRGE), a lightweight SwiGLU-gated probing architecture designed to suppress schema noise and amplify localized question--schema mismatch signals that indicate unanswerability.
    \item \textbf{Empirical validation and analysis.} We evaluate \textsc{LatentRefusal} on diverse unanswerable and ambiguous Text-to-SQL settings, demonstrating improved refusal reliability at near-instruction-following cost, and provide ablations and interpretability analyses that isolate where refusal signals emerge in the latent space.
\end{itemize}

\section{Related Work}
\label{sec:related_work}

\paragraph{Unanswerability and ambiguity in Text-to-SQL.}
Real-world Text-to-SQL must handle \emph{ambiguous} questions (multiple valid interpretations) and \emph{unanswerable} questions (cannot be grounded to the available schema/database). Prior work studies intention types and fine-grained unanswerability categories \citep{zhang2020did,wang2022know}, extends the setting to multi-turn conversations \citep{dong2024practiq}, and benchmarks linguistic ambiguity with paired ambiguous/clarified inputs \citep{saparina2024ambrosia}. In this paper, we target a stricter system requirement: a \textbf{low-latency, pre-generation gate} that decides whether \emph{any} SQL should be produced.

\paragraph{Prompt-based refusal and output-based uncertainty.}
Prompting an LLM to abstain is convenient and training-free, but refusal is prompt-sensitive and can fail exactly when the model hallucinates, after the system has already entered executable-code decoding. Output-based uncertainty uses sampling and disagreement signals (e.g., SelfCheckGPT and semantic entropy) \cite{manakul2023selfcheckgpt,farquhar2024detecting,kuhn2023semantic,wang2023selfconsistency}, while self-evaluation and selective generation use single-pass confidence, OOD-style scores, or verbalized uncertainty \cite{kadavath2022language,ren2023ood,lin2022teaching,xiong2024can}. These methods are broadly applicable but often conflict with Text-to-SQL deployment constraints: multi-sampling is costly, and comparing SQL candidates can require execution or expensive reasoning, which is misaligned with \textbf{pre-execution} safety gating.

\paragraph{Latent/internal-state reliability signals.}
Recent work shows that reliability and truthfulness are reflected in latent space and internal activations, including latent truth directions (CCS) \citep{burns2023discovering,marks2024geometry}, generation-driven hallucination membership estimation (HaloScope) \citep{du2024haloscope}, and internal-state predictors of hallucination \citep{azaria_mitchell2023lying,chen2024inside}. These approaches motivate \textbf{single-pass, pre-generation} decisions when internal states are accessible. Our work follows this line but addresses a Text-to-SQL-specific challenge: schema-conditioned prompts are dominated by long schema tokens while refusal cues are sparse and localized, motivating a probe that suppresses schema noise and amplifies question-schema mismatch evidence \cite{kossen2024semantic}.

\section{\textsc{LatentRefusal}}
\label{sec:latentrefuse}

\begin{figure*}[t]
    \centering
    \includegraphics[width=\linewidth]{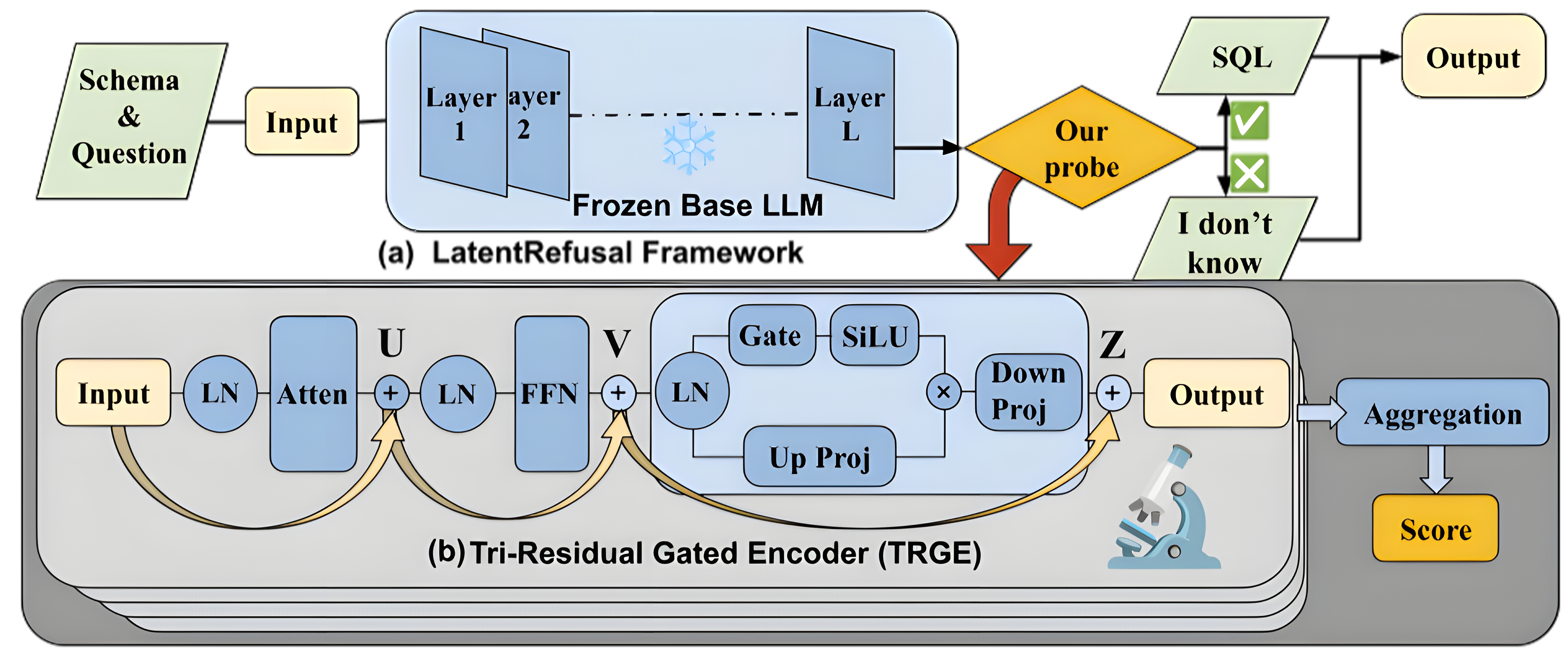}
    \caption{Overview of the \textsc{LatentRefusal} framework. \textbf{(a) Refusal gating pipeline:} given a natural-language question and a database schema, a frozen base LLM (e.g., Qwen-3-8B or Llama-3.1-8B) produces hidden states at a selected intermediate layer $l^\star$; a lightweight TRGE probe predicts answerability probability $\hat{p}$ \emph{before} any SQL tokens are generated, and a binary gate (threshold $\tau$) either triggers SQL generation or returns a safe refusal response. \textbf{(b) TRGE probe architecture:} each Tri-Residual Gated Encoder layer augments a standard Transformer block (self-attention + FFN) with a third SwiGLU-gated residual branch, designed to suppress schema noise and amplify sparse, localized question--schema mismatch cues indicative of unanswerability.}
    \label{fig:framework_arch}
    \vspace{-1em}
\end{figure*}

\textsc{LatentRefusal} is a \emph{latent-signal} refusal mechanism for Text-to-SQL that decides whether to answer \emph{prior to generation} by reading refusal cues from a frozen LLM's internal representations (Figure~\ref{fig:framework_arch}). The core novelty is to treat refusal as a \emph{representation-level} decision problem rather than an \emph{output-level} instruction-following behavior: instead of asking the LLM to say ``I don't know'' (which can fail under hallucination), we learn a compact detector on intermediate hidden states and insert it as a deterministic gate.

\subsection{Refusal Gating Framework}
\label{subsec:gating_framework}

\paragraph{Why a pre-generation gate?}
Text-to-SQL differs from open-ended generation in that the model output is executable code. When the query is unanswerable or underspecified (e.g., missing columns/tables, ambiguous constraints, non-existent entities, or out-of-scope requests), ``helpful'' decoding can still yield plausible SQL that executes successfully but answers the \emph{wrong} question. Existing approaches face two limitations:
(i) \textbf{Prompt-based refusal} relies on the model's decoded text to abstain, which is prompt-sensitive and may collapse precisely when the model enters a hallucination state.
(ii) \textbf{Uncertainty-based refusal} often requires multi-sampling; in Text-to-SQL, comparing candidate programs can require execution or expensive equivalence checking, which conflicts with safety gating and increases latency.

\paragraph{Design goal.}
We aim for a refusal mechanism that is (i) \textbf{single-pass} (one forward pass through the base LLM), (ii) \textbf{pre-generation} (no SQL tokens are generated unless permitted), (iii) \textbf{execution-free} (no database interaction during gating), and (iv) \textbf{architecture-independent} (applicable to any frozen LLM with accessible intermediate activations). Concretely, ``architecture-independent'' means we do not update the base LLM parameters, while the probe is trained with lightweight supervision. The method requires white-box access to intermediate hidden states, which aligns with our target application (privacy-sensitive financial QA) where models are typically deployed locally rather than via closed APIs.

\paragraph{Gating rule.}
We implement a deterministic gate that intercepts the generation process based on the probe's output. Specifically, the system returns a refusal response if the predicted answerability probability $\hat{p}=g_\phi(\mathbf{H})$ falls below a calibrated threshold $\tau$ (i.e., $\hat{p}<\tau$), and proceeds to SQL generation otherwise. This mechanism decouples safety judgment from generation, ensuring unanswerable queries are blocked before any potentially harmful SQL is produced. This setup is consistent with the \textbf{reject option} and \textbf{selective classification} frameworks \cite{chow1970optimum,el2010foundations,geifman2017selective,geifman2019selectivenet}, where a model abstains from answering under uncertainty to achieve a better safety--utility trade-off.

\paragraph{What signal do we use?}
A frozen LLM encodes rich consistency information between question and schema during the forward pass, even when its final generation may hallucinate. \textsc{LatentRefusal} exploits this by operating directly on internal hidden states rather than on decoded tokens. Concretely, let $\mathcal{M}$ be the frozen base LLM and let $\mathbf{H}^{(l)} \in \mathbb{R}^{T \times d}$ denote the hidden states at layer $l$. We choose one layer index $l^\star$ (validated on development data) and feed $\mathbf{H}^{(l^\star)}$ to the probe:
\begin{equation}
\label{eq:probe_pred}
\hat{p} = g_\phi\!\left(\mathbf{H}^{(l^\star)}\right),
\qquad
\mathbf{H}^{(l^\star)} = \mathcal{M}^{(l^\star)}(x).
\end{equation}
where $\phi$ represents the trainable parameters of the probe, and $x$ is the input sequence.
Using a single intermediate layer is intentionally minimal: it keeps the gate lightweight and avoids entanglement with the base model's decoder dynamics.

\subsection{TRGE Probe}
\label{subsec:trge}

\paragraph{Challenge: schema-heavy prompts hide sparse refusal cues.}
In Text-to-SQL prompting, the schema tokens frequently dominate context length. However, answerability evidence is often localized: a single missing column mention, an unmatched entity, or an impossible join path. Simple pooling with a linear classifier can underfit because the relevant signal is sparse and can be overwhelmed by irrelevant schema content. A vanilla Transformer probe helps, but without an inductive bias to \emph{suppress} schema noise it may still spend capacity modeling schema regularities rather than detecting mismatches.

\paragraph{Our probe design.}
We propose the \textbf{Tri-Residual Gated Encoder (TRGE)}, a small Transformer-style encoder that adds a \emph{third} residual branch consisting of a SwiGLU gating module. The motivation is explicit: a content-aware gate can act as a soft feature selector that down-weights irrelevant schema patterns while amplifying mismatch features correlated with unanswerability.

Given an input representation $\mathbf{Z}^{(k-1)} \in \mathbb{R}^{T \times d}$ to TRGE layer $k$ (where $d=512$), we compute:
\begin{equation}
\label{eq:trge_computation}
\begin{aligned}
\mathbf{U} &= \mathbf{Z}^{(k-1)} + \operatorname{Attn}\!\left(\operatorname{LN}(\mathbf{Z}^{(k-1)})\right),\\
\mathbf{V} &= \mathbf{U} + \operatorname{MLP}\!\left(\operatorname{LN}(\mathbf{U})\right),\\
\mathbf{Z}^{(k)} &= \mathbf{V} + \operatorname{SwiGLU}\!\left(\operatorname{LN}(\mathbf{V})\right),
\end{aligned}
\end{equation}
where $\operatorname{LN}$ is layer normalization and $\operatorname{Attn}$ is multi-head self-attention with 8 heads. Here, $\mathbf{U}$ and $\mathbf{V}$ denote the intermediate residual states after the attention and MLP blocks, respectively. The gating branch utilizes the SwiGLU activation:
\begin{equation}
\label{eq:swiglu}
\operatorname{SwiGLU}(\mathbf{x})
=
\mathbf{W}_d\Big(\operatorname{SiLU}(\mathbf{W}_g\mathbf{x}) \odot (\mathbf{W}_u\mathbf{x})\Big),
\end{equation}
{\sloppy
with learnable matrices $\mathbf{W}_g,\mathbf{W}_u,\mathbf{W}_d$ and ele\-ment-wise product $\odot$. Here, $\mathbf{x}$ is the input vector, and $\operatorname{SiLU}$ is the Sigmoid Linear Unit ac\-ti\-va\-tion func\-tion.\par}

\paragraph{Why tri-residual gating helps.}
The extra residual branch provides a direct pathway for ``refusal-relevant'' features to accumulate across layers without being washed out by attention/MLP mixing. Intuitively, the gate
$
\mathbf{g}=\operatorname{SiLU}(\mathbf{W}_g\mathbf{x})
$
acts as a soft mask that is \emph{input-conditioned}: it can suppress schema boilerplate while preserving tokens/positions carrying mismatch evidence. This is particularly suited to schema-conditioned prompts where the majority of tokens are irrelevant to the answerability decision.

After $L_p$ TRGE layers, we aggregate token representations into a single vector and predict a scalar score (Eq. \ref{eq:score_pred}):
\begin{equation}
\label{eq:score_pred}
\mathbf{r}=\operatorname{Agg}\!\left(\mathbf{Z}^{(L_p)}\right), \qquad s = \mathbf{w}^\top \mathbf{r} + b, \qquad \hat{p}=\sigma(s).
\end{equation}
{\sloppy
where $\mathbf{r}$ is the pooled rep\-re\-sen\-ta\-tion, $\mathbf{w}$ and $b$ are the linear clas\-si\-fier's weight and bias, $s$ is the scalar logit, and $\sigma$ denotes the sigmoid function.
We use a simple ag\-gre\-ga\-tion op\-er\-a\-tor $\operatorname{Agg}(\cdot)$ (mean pooling unless otherwise noted), keeping the probe light\-weight; Section~\ref{sec:experiments} studies al\-ter\-na\-tives.\par}

\subsection{Training and Implementation Details}
\label{subsec:training}

We freeze $\mathcal{M}$ entirely and train only the probe parameters $\phi$. This yields three practical benefits: (i) stable behavior of the underlying generator, (ii) minimal additional compute and memory, and (iii) easy attachment to different base LLMs.

\subsubsection{Supervision on hidden states}
Given a labeled dataset $\mathcal{D}=\{(S_i,Q_i,y_i)\}_{i=1}^N$ where $y_i\in\{0,1\}$ indicates whether the query is answerable under schema $S_i$, we build the model input via a template $\mathcal{T}$:
\begin{equation}
\label{eq:input_template}
\mathbf{x}_i=\mathcal{T}(S_i,Q_i).
\end{equation}
where $\mathbf{x}_i$ is the tokenized input sequence corresponding to schema $S_i$ and question $Q_i$.
We run a single forward pass through the frozen LLM and extract hidden states from layer $l^\star$:
\begin{equation}
\label{eq:hidden_extract_train}
\mathbf{H}_i = \mathbf{H}_i^{(l^\star)}=\mathcal{M}^{(l^\star)}(\mathbf{x}_i)\in\mathbb{R}^{T_i\times d}.
\end{equation}
where $T_i$ denotes the sequence length of the $i$-th sample.
The probe predicts $\hat{p}_i=g_\phi(\mathbf{H}_i)$ and is trained on $\{(\mathbf{H}_i,y_i)\}_{i=1}^N$.

\subsubsection{Numerical stability}
During mixed-precision inference or offline hidden-state extraction, rare NaN values can destabilize optimization. We apply sanitization operator $\Psi(\cdot)$:
\begin{equation}
\label{eq:sanitization}
\tilde{\mathbf{H}}_i=\Psi(\mathbf{H}_i), \quad
\Psi(h)=
\begin{cases}
0,& h\ \text{is NaN},\\
\ \ \ C,& h=+\infty,\\
- C,& h=-\infty,\\
h,& \text{otherwise},
\end{cases}
\end{equation}
with a large clipping constant $C$ (e.g., $10^4$), followed by token-wise normalization:
\begin{equation}
\label{eq:normalization}
\mathbf{H}^{\text{safe}}_i=\operatorname{LN}(\tilde{\mathbf{H}}_i).
\end{equation}
where $\tilde{\mathbf{H}}_i$ is the sanitized hidden state matrix, and $\mathbf{H}^{\text{safe}}_i$ denotes the final stabilized input for the probe.
We feed $\mathbf{H}^{\text{safe}}_i$ to the probe in both training and evaluation for consistent behavior.

\subsubsection{Optimization objective and thresholding}
Let $s_i=f_\phi(\mathbf{H}^{\text{safe}}_i)$ be the probe logit and $\hat{p}_i=\sigma(s_i)$. We optimize binary cross-entropy:
\begin{equation}
\label{eq:loss}
\mathcal{L}(\phi)
=-\frac{1}{N}\sum_{i=1}^{N}\Big(y_i\log\hat{p}_i+(1-y_i)\log(1-\hat{p}_i)\Big).
\end{equation}
where $N$ is the number of samples in the training set.
For a single example, the derivative w.r.t.\ the logit is:
\begin{equation}
\label{eq:derivative}
\frac{\partial \ell}{\partial s}=\sigma(s)-y.
\end{equation}
where $\ell$ represents the loss for a single instance.
At deployment, we select the threshold $\tau$ on a development set to satisfy a desired safety--utility operating point (e.g., high refusal recall under a bounded false-refusal rate). This yields an explicit, auditable trade-off suitable for production gating.

\textsc{LatentRefusal} is novel in (i) making refusal a \emph{pre-generation} decision using \emph{latent} signals from a frozen LLM, enabling single-pass, execution-free gating; and (ii) introducing TRGE, a tri-residual SwiGLU-gated probe tailored to schema-heavy Text-to-SQL prompts, designed to suppress schema noise and amplify sparse mismatch cues that drive unanswerability.

\section{Experiments}\label{sec:experiments}

We evaluate our TRGE Transformer probe for detecting unanswerability across four datasets and compare against representative baselines.

\subsection{Experimental Setup}

\paragraph{Datasets.} We evaluate on four benchmarks: (1) \textbf{TriageSQL} \cite{zhang2020did}: converted into a binary refusal task focusing on medical intent; (2) \textbf{AMBROSIA} \cite{saparina2024ambrosia}: testing sensitivity to linguistic ambiguity via paired clear vs. ambiguous questions; (3) \textbf{SQuAD 2.0} \cite{rajpurkar2018know}: used to evaluate cross-task generalizability from Text-to-SQL to machine reading comprehension; and (4) \textbf{MD-Enterprise}\footnote{This is an internal dataset and cannot be released due to privacy reasons.}: a Chinese vertical industrial benchmark spanning six domains (Stock, HR, etc.) with expert-annotated answerability labels based on business logic and safety constraints.

\paragraph{Models and Inference.} We use Qwen-3-8B and Llama-3.1-8B \cite{yang2025qwen3,grattafiori2024llama} as local backbones, loaded in bfloat16. For sampling-based baselines (Semantic Entropy, Eigenscore), we sample Top-$K=10$ outputs at temperature $T=0.7$; single-pass baselines (Self-evaluation, CCS, TSV, HaloScope, SAPLMA) use greedy decoding. For \textsc{LatentRefusal}, we extract hidden states from a single greedy forward pass ($T=0$). The probe is trained with a learning rate of $10^{-5}$, batch size 8, no warm-up, and a fixed random seed of 42.

\paragraph{Training Protocol and Splits.}
We train one probe per dataset group. When multiple datasets are available within a single deployment, we merge them and train jointly. For instance, our internal MD-Enterprise dataset spans six domains (Stock, HR, Loan, Retail, Risk Control, Supervise), and joint training performs well across all of them; AMBROSIA likewise includes multiple ambiguity types and is trained jointly. In Table~\ref{tab:main_results}, each dataset is trained and evaluated separately. We split training/validation as 8:2 and sample 300 examples as the held-out test set for each dataset.

\begin{table*}[t]
\centering
\resizebox{\textwidth}{!}{%
\begin{tabular}{llccccc}
\toprule
\textbf{LLM} & \textbf{Method} & \textbf{MD-Enterprise} & \textbf{AMBROSIA} & \textbf{SQuAD} & \textbf{TriageSQL} & \textbf{Avg.F1} \\
\midrule
\multirow{8}{*}{\textbf{Llama-3.1-8B}} 
& Semantic Entropy & 66.1 & 62.1 & 82.3 & 66.7 & 69.3 \\
& CCS & 53.1 & 54.1 & 62.6 & 62.9 & 58.2 \\
& Self-evaluation* & 66.7 & 64.6 & 74.2 & 67.2 & 68.2 \\
& Eigenscore & 82.5 & 63.2 & 72.4 & 77.8 & 73.8 \\
& TSV & 97.4 & 74.3 & 74.7 & 85.2 & 82.9 \\
& HaloScope & 97.0 & 73.7 & 66.1 & 82.8 & 79.9 \\
& SAPLMA* & 97.5 & 77.7 & 75.4 & 81.0 & 82.9 \\
& \cellcolor{gray!25}\textbf{\textsc{LatentRefusal}} & \cellcolor{gray!25}\textbf{99.6} & \cellcolor{gray!25}\textbf{80.2} & \cellcolor{gray!25}\textbf{86.6} & \cellcolor{gray!25}\textbf{87.7} & \cellcolor{gray!25}\textbf{88.5} \\
\midrule
\multirow{8}{*}{\textbf{Qwen-3-8B}} 
& Semantic Entropy & 72.7 & 58.0 & 82.4 & 66.6 & 70.0 \\
& CCS & 55.4 & 47.0 & 82.3 & 73.4 & 64.5 \\
& Self-evaluation* & 68.1 & 60.2 & 87.4 & 56.3 & 68.0 \\
& Eigenscore & 80.0 & 59.1 & 70.0 & 77.8 & 71.7 \\
& TSV & 98.9 & 73.3 & 78.0 & 85.0 & 83.8 \\
& HaloScope & 98.0 & 72.8 & 80.1 & 80.0 & 82.7 \\
& SAPLMA* & 97.8 & 81.2 & 76.8 & 82.6 & 84.6 \\
& \cellcolor{gray!25}\textbf{\textsc{LatentRefusal}} & \cellcolor{gray!25}\textbf{98.8} & \cellcolor{gray!25}\textbf{80.9} & \cellcolor{gray!25}\textbf{88.6} & \cellcolor{gray!25}\textbf{87.1} & \cellcolor{gray!25}\textbf{88.8} \\
\midrule
\textbf{DeepSeek-Chat (API)} & Prompt-based & 97.2 & 70.3 & 87.4 & 77.8 & 83.2 \\
\bottomrule
\end{tabular}
}
\caption{Refusal detection performance (F1 \%) across four benchmarks: MD-Enterprise (Chinese financial QA, 6 domains), AMBROSIA (linguistic ambiguity), SQuAD 2.0 (cross-task reading comprehension), and TriageSQL (medical intent). Local backbone models (Llama-3.1-8B and Qwen-3-8B) are loaded in bfloat16 with max sequence length 2048; DeepSeek-Chat is accessed via API. Sampling-based baselines (Semantic Entropy, Eigenscore) use $K{=}10$ samples at $T{=}0.7$; Self-evaluation uses a single-pass logit score; \textsc{LatentRefusal} uses a single greedy forward pass. Best results per backbone are in \textbf{bold}. Methods marked with * are reproduced by us; all others use official implementations.}
\label{tab:main_results}\vspace{-1em}\end{table*}

\paragraph{Baselines.} We compare against three categories of methods:
(1) \textbf{Output-based Uncertainty}: \textit{Self-evaluation} \citep{kadavath2022language} prompts the model to estimate its own correctness; \textit{Semantic Entropy} \cite{farquhar2024detecting} measures uncertainty via agreement across sampled generations.
(2) \textbf{Internal-State Methods}: We evaluate unsupervised approaches \textit{CCS} \citep{burns2023discovering}, \textit{Eigenscore} \cite{chen2024inside}, and weakly-supervised \textit{HaloScope} \citep{du2024haloscope}, alongside supervised probes \textit{SAPLMA} \citep{azaria_mitchell2023lying} and \textit{TSV} \cite{park2025steer} which predict truthfulness from hidden states or steering vectors.
(3) \textbf{Prompting}: \textit{DeepSeek-Chat} serves as a zero-shot instruction-following baseline.

\subsection{Main Results}

Table~\ref{tab:main_results} summarizes the refusal detection performance across four benchmarks. We report F1 as the primary metric, computed at a fixed decision threshold tuned on development data.
\paragraph{Overall performance.}
{\sloppy
\textsc{LatentRefusal} achieves the highest average F1 on both backbone models: \textbf{88.5\%} on Llama-3.1-8B and \textbf{88.8\%} on Qwen-3-8B, out\-per\-forming all baselines by sub\-stan\-tial margins. Specifically, it surpasses the strongest baseline---SAPLMA (84.6\% on Qwen-3-8B) and TSV/SAPLMA (82.9\% on Llama-3.1-8B)---by +4.2 and +5.6 points re\-spec\-tive\-ly. Notably, against the internal-state baseline Eigenscore, the gains exceed +14 points, dem\-on\-strat\-ing that the TRGE probe ar\-chi\-tec\-ture more ef\-fec\-tive\-ly distills refusal-relevant signals from hidden states than un\-su\-per\-vised spectral methods.\par}
\paragraph{Robustness to Linguistic Ambiguity.}
The AMBROSIA dataset evaluates sensitivity to underspecified queries with subtle linguistic ambiguity. Here, \textsc{LatentRefusal} achieves \textbf{80.2\%} (Llama), outperforming Semantic Entropy by \textbf{18.1 points}. This significant gap may partly stem from a known challenge of sampling-based uncertainty in Text-to-SQL, where syntactic diversity can mask semantic overconfidence. In our implementation, we approximate equivalence using syntactic agreement (Appendix C.1) for safety and cost reasons, which may underestimate the best-case performance of semantic clustering. Our latent-signal approach avoids this limitation by detecting the underlying representation-level confusion before it manifests as overconfident decoding.

\vspace{-0.7em}
\paragraph{Training Efficiency.}
Instead of relying on zero-shot transfer, \textsc{LatentRefusal} attains its high performance through highly efficient supervised adaptation. Our method can be trained on any dataset using only $\sim$300 samples, completing in just 10 minutes on a single A100-80G GPU. This efficiency allows custom refusal gates to be deployed rapidly for new domains (e.g., reaching \textbf{88.6\%} F1 on SQuAD 2.0 and \textbf{87.1\%} F1 on TriageSQL) with minimal data annotation and computational cost.

\vspace{-0.6em}
\paragraph{Comparison with API-based prompting.}
The DeepSeek-Chat API baseline uses zero-shot prompting to elicit refusal. While it achieves reasonable performance (83.2\% avg.), it underperforms \textsc{LatentRefusal} by 5+ points and lacks controllability---prompt-based refusal is sensitive to instruction phrasing and can fail precisely when the model hallucinates. Our internal-signal approach provides a more reliable and deterministic safety gate.

\subsection{Efficiency Analysis}
\label{subsec:efficiency}

Table~\ref{tab:efficiency} compares inference latency (Qwen-3-8B backbone). \textsc{LatentRefusal} adds negligible overhead (\textbf{2ms}), achieving 54ms total latency---\textbf{13.7$\times$ faster} than Semantic Entropy (740ms). While sampling methods are prohibitive for real-time use, \textsc{LatentRefusal} achieves a favorable accuracy–latency trade-off and lies on the Pareto frontier among the evaluated baselines (Table~\ref{tab:main_results}). Specifically, it attains the highest F1 (88.8\%) at near-minimal latency, superior to both computationally expensive spectral methods and similarly fast but less accurate supervised baselines (TSV).

\begin{table}[ht]
\centering
\resizebox{\columnwidth}{!}{%
\begin{tabular}{lccc}
\toprule
\textbf{Method} & \textbf{\# Runs} & \textbf{Main Latency(ms)} \\
\midrule
Semantic Entropy & N=10 & 52*10 \\
CCS & 2 & 100 \\
Self-evaluation* & 1 & 53 \\
Eigenscore & $N=5/10$ & $51.7*N$ \\
TSV & 1 & 52 \\
HaloScope & 1 & 50 \\
SAPLMA* & 1 & 53 \\
\textsc{LatentRefusal} & 1 & \ 54 \\
\bottomrule
\end{tabular}
}
\caption{Inference latency comparison on the Qwen-3-8B backbone (single A100-80G GPU, bfloat16, sequence length 2048). ``\# Runs'' denotes the number of LLM forward passes required. Sampling-based methods (Semantic Entropy, Eigenscore) scale linearly with the number of samples $N$. \textsc{LatentRefusal} requires only one forward pass plus a 2ms probe, achieving the best accuracy--latency trade-off on the Pareto frontier.}
\label{tab:efficiency}

\end{table}

\paragraph{Architectural Efficiency.}
Efficiency stems from three factors: (1) \textbf{Zero-redundancy extraction}: reusing mandatory forward-pass states adds no LLM-level compute; (2) \textbf{Parameter efficiency}: the 19M-parameter TRGE probe is $<0.3\%$ of the backbone size; (3) \textbf{Early-exit capability}: deciding refusal \emph{before} decoding saves generation costs for unanswerable queries.

\subsection{Analysis}

\paragraph{The Failure of Spectral Uncertainty.}
Eigenscore relies on spectral statistics of hidden states across multiple samples. However, in Text-to-SQL, "confident hallucinations" are common: the model may generate syntactically diverse SQL (e.g., varying `JOIN` orders or alias names) that are semantically identical. This diversity inflates eigenvalue spread, causing spectral methods to misinterpret syntactic variance as epistemic uncertainty. TRGE avoids this by learning to identify the \emph{source} of mismatch in the prompt-schema representation, which is invariant to the downstream decoding path.

\paragraph{Cross-Backbone Consistency.}
The comparable performance between Llama-3.1-8B (88.5\%) and Qwen-3-8B (88.8\%) is consistent with the hypothesis that refusal signals may be encoded in a structurally similar manner across modern Transformer architectures. This suggests that the TRGE architecture is not overfitted to a specific model's quirks, and implies that the probe may capture a task-general signal regarding how LLMs process unanswerable context.

\paragraph{Error Analysis and Future Work.}
Qualitative inspection reveals two primary failure modes: (1) \textbf{Semantic Near-Misses}: when a column name is semantically similar but logically incorrect (e.g., `revenue` vs. `gross\_profit`), the probe occasionally underestimates uncertainty. (2) \textbf{Deep Reasoning Chains}: queries requiring complex multi-hop joins sometimes exhibit weak mismatch signals in the selected layer. Future work could explore \emph{multi-layer fusion} or \emph{schema-aware attention} to better capture these high-order logical inconsistencies.

\subsection{Ablation Studies}
\label{subsec:ablation}

We conduct comprehensive ablation studies on Qwen-3-8B using the TriageSQL dataset to validate the architectural decisions of \textsc{LatentRefusal}. Additional ablations on probe depth and training strategies are provided in Appendix~\ref{sec:appendix_ablation}.

\subsubsection{Architecture and Layer Selection}

We validate the TRGE probe design and investigate the optimal source of refusal signals.

\paragraph{Efficacy of the TRGE Architecture.}
{\sloppy
Table \ref{tab:arch_ablation} isolates the impact of the Tri-Residual Gated Encoder. The full TRGE model achieves an F1 score of \textbf{87.1\%}. Removing the gating branch (\textit{w/o SwiGLU}) degrades per\-for\-mance to 85.4\%, but the most critical insight comes from replacing the SwiGLU gate with a standard MLP ($-$4.1\% F1) or a Linear Probe ($-$16.7\% F1). The failure of the Linear Probe (70.4\% F1) confirms that refusal de\-tec\-tion requires modeling complex non-linear in\-ter\-ac\-tions between question and schema. Moreover, replacing SwiGLU with a standard MLP drops per\-for\-mance to 83.0\%, dem\-on\-strat\-ing that the \textit{gating mech\-a\-nism}---which se\-lec\-tive\-ly suppresses schema noise---is essential. Among gating variants, SwiGLU out\-per\-forms GLU and GeGLU, likely due to its smoother op\-ti\-mi\-za\-tion land\-scape.\par}

\begin{table}[t]
\centering
\begin{tabular}{lcc}
\toprule
\textbf{Variant} & \textbf{F1 (\%)}  & \textbf{Time (ms)} \\
\midrule
\rowcolor{gray!25}
TRGE (Full) & \textbf{87.1} & 2.6 \\
w/o SwiGLU & 85.4 & 2.3 \\
SwiGLU $\rightarrow$ MLP & 83.0 & 2.2 \\
SwiGLU $\rightarrow$ GLU & 75.5 & 2.3 \\
SwiGLU $\rightarrow$ GeGLU & 85.1 & 2.0 \\
Linear Probe & 70.4 & \textbf{0.8} \\
\bottomrule
\end{tabular}
\caption{Architecture ablation on Qwen-3-8B / TriageSQL (layer $-16$, 4 probe layers, dropout 0.2). We compare the full TRGE against five variants: removing the SwiGLU branch, replacing it with MLP/GLU/GeGLU gates, and a linear probe baseline. ``Time'' reports the probe-only latency (excluding LLM forward pass). The SwiGLU gate contributes +16.7\% F1 over the linear probe and +4.1\% over the MLP replacement.}
\label{tab:arch_ablation}
\end{table}

\begin{table}[t]
\centering
\setlength{\tabcolsep}{3pt}
\begin{tabular}{lccccc}
\toprule
\textbf{Layer} & \textbf{Acc} & \textbf{Prec} & \textbf{Rec} & \textbf{AUC} & \textbf{F1} \\
\midrule
$-1$ & 84.4 & 76.7 & 99.0 & 87.6 & 86.5 \\
$-8$ & 84.5 & \textbf{77.4} & 97.8 & \textbf{88.7} & 86.4 \\
\rowcolor{gray!25}
$-\textbf{16}$ & \textbf{85.0} & 77.2 & \textbf{99.8} & 88.4 & \textbf{87.1} \\
$-24$ & 84.4 & 76.4 & \textbf{99.8} & 87.7 & 86.5 \\
$-32$ & 82.9 & 74.7 & \textbf{99.8} & 88.4 & 85.4 \\
\bottomrule
\end{tabular}
\caption{Hidden state layer selection ablation on Qwen-3-8B / TriageSQL. Layer indices are relative to the final layer ($-1$ = last, $-32$ = first). The TRGE probe (4 layers, dropout 0.2) is trained separately for each source layer. Layer $-16$ (middle of the 32-layer backbone) achieves the best F1 (87.1\%) and near-perfect recall (99.8\%), supporting the hypothesis that refusal signals peak at intermediate layers before being collapsed by later processing.}
\label{tab:layer_ablation}

\end{table}

\paragraph{Locating Refusal Signals.}
Table \ref{tab:layer_ablation} shows that the optimal refusal signal resides in the middle-to-late layers (Layer $-16$), achieving the highest Accuracy (85.0\%) and F1 (87.1\%). While Layer $-8$ offers slightly higher precision, Layer $-16$ provides a superior balance with near-perfect recall (99.8\%). Notably, the final layer ($-1$) exhibits lower recall (99.0\%) than deeper layers. This finding supports the "mechanistic interpretability" hypothesis that LLMs encode epistemic uncertainty about unanswerable queries in intermediate processing stages, which may be resolved—or collapsed into confident hallucinations—by the time representations reach the final output layer \cite{marks2024geometry,kossen2024semantic}.
\paragraph{Probe Depth and Capacity.}
Table \ref{tab:depth_ablation} reveals an inverted-U relationship between probe depth and detection accuracy. Performance peaks at 4 layers (87.1\% F1). Shallower probes (1--2 layers) underfit the data, while deeper probes (8--12 layers) show diminishing returns and signs of overfitting. This suggests that a compact 4-layer probe is sufficient to extract the refusal signal without memorizing dataset-specific schema artifacts, validating our design goal of a lightweight, low-latency module.
\begin{table}[ht]
\centering
\begin{tabular}{cccc}
\toprule
\textbf{Layers} & \textbf{Params} & \textbf{F1 (\%)} & \textbf{Time (ms)} \\
\midrule
1 & 9.6M & 82.03 & 1.04 \\
2 & 12.7M & 83.87 & 1.64 \\
\rowcolor{gray!25}
4 & 19.0M & \textbf{87.09} & 2.60 \\
6 & 25.3M & 85.28 & 3.40 \\
8 & 31.7M & 84.61 & 4.45 \\
12 & 44.3M & 83.02 & 6.46 \\
\bottomrule
\end{tabular}
\caption{Probe depth ablation on Qwen-3-8B / TriageSQL (layer $-16$, SwiGLU gate, dropout 0.2). We vary the number of TRGE layers from 1 to 12 and report F1 and probe-only inference time. Performance peaks at 4 layers (19.0M params, 2.60ms), with deeper probes showing diminishing returns due to overfitting on the 300-sample training set.}
\label{tab:depth_ablation}
\end{table}
\paragraph{Robustness to Training Strategy.}
Our method is robust to hyperparameter variations (Table \ref{tab:loss_ablation} and \ref{tab:dropout_ablation} in Appendix). Label Smoothing ($\epsilon=0.1$) achieves optimal performance (87.1\% F1) by mitigating overconfidence. Additionally, stability across varying dropout rates (0.0--0.3) indicates the probe learns distributed features rather than brittle artifacts.
\paragraph{Summary of ablations.}
Our ablation studies confirm that: (1) the SwiGLU gating mechanism is critical, with a 16.7\% F1 gap between TRGE and linear probes; (2) intermediate LLM layers provide the best refusal signals; and (3) training is robust to standard hyperparameter variations.
\section{Conclusion}
We address the critical safety risk of unanswerable queries in Text-to-SQL, where hallucinations yield harmful executable code. We formalize refusal as a strict \emph{pre-generation gating} problem and propose \textsc{LatentRefusal}, a single-pass mechanism that detects answerability directly from a frozen LLM's internal states. By introducing the Tri-Residual Gated Encoder (TRGE), we effectively amplify sparse refusal signals amidst schema noise, avoiding the latency and risks of execution-based uncertainty estimation. Empirical results confirm that \textsc{LatentRefusal} provides a robust, low-latency safety layer essential for production-grade Text-to-SQL systems.

\section*{Limitations}
While \textsc{LatentRefusal} achieves high detection accuracy, its generalization capability across disparate domains remains a limitation; the probe currently requires fine-tuning to adapt to specific deployment scenarios. However, this need for adaptation is offset by the method's exceptional training efficiency. The probe introduces minimal computational overhead, with training converging in approximately 10 minutes on a single NVIDIA A100-80G GPU. This low-cost training profile makes it practical to re-train or fine-tune the refusal mechanism for new verticals without significant resource expenditure. Future research will explore cross-dataset generalization and universal safety gating—where a single probe can reliably gate any Text-to-SQL deployment without domain-specific re-tuning—through techniques such as domain-invariant representation learning and meta-learning.

\nocite{sahoo2024systematic,qin2022survey,cheng2024can,zou2023representation,manakul2023selfcheckgpt,park2025steer,wang2022know,farquhar2024detecting,chen2024inside,vaswani2017attention,yang2025qwen3,grattafiori2024llama,zhang2020did,saparina2024ambrosia,rajpurkar2016squad,tsai2020evaluating}

\section*{Ethics Statement}
This work does not involve human subjects or personally identifiable information. The MD-Enterprise dataset is used under institutional agreement and cannot be released due to privacy constraints. All other datasets are publicly available benchmarks.

\section*{Acknowledgements}
This work was supported by the Yangtze River Delta Science and Technology Innovation Community Joint Research Project (YDZX20233100004031), the ``Internet Financial Customer Service Vertical Large Model Construction Project'' (Contract Registration No. 2026310031000213), and the Gientech Enterprise Cooperation Project on Financial AI Product Safety Monitoring and Intelligent Processing of Massive Transaction Data (GIFD20222025).

\bibliography{custom}

\clearpage
\onecolumn
\appendix

\begin{center}
{\Large\bfseries Appendix}
\end{center}
\vspace{0.5em}

\section*{Appendix Contents}
\makeatletter
\app@starttoc
\clearpage
\immediate\openout\app@toc=\jobname.apptoc
\app@tocwritetrue
\makeatother

\section{Prompt Template}
\label{sec:prompt_template}

We include the prompt templates used for answerability judgment (Chinese and English).

\begin{CJK*}{UTF8}{gbsn}
\begin{tcolorbox}[breakable, colback=gray!10, colframe=blue!40, title=Prompt Template (Chinese)]
现在是2025年，你是一个会写sql的机器，你不具备数据分析能力。
不要对开放问题给出你的看法。

你的任务是以最严格的标准判断：用户的问题是否为SQL查询需求，
是否有确定的计算逻辑，能否通过生成并执行sql计算，在给定的
数据库范围内，得出确定的答案。

\textcolor{magenta}{\{database\_meta\}}

\#\#\# User Query:

\textcolor{magenta}{\{user\_query\}}

\#\#\# 要求：仔细分析用户输入和数据库信息，按照以下例子进行回答。

可以回答的问题具备以下特征：

1. 查询具体的历史数据点，如某个时间点的价格、交易量等\\
2. 对历史数据进行简单的统计计算，如平均值、总和、最大/最小值等\\
3. 查找满足特定条件的记录，如超过某个阈值的数据\\
4. 查询实体的基本属性或标识信息，如代码、名称等\\
5. 在特定时间范围内进行以上操作

不能回答的问题特征：

1. 没有具体、直接的计算逻辑的。比如：如何分析员工的职位晋升？\\
2. 涉及未来预测或趋势判断，比如：未来员工会不会离职？\\
3. 需要主观分析或评估，比如：员工的工作能力如何？\\
4. 需要数据库之外的其他信息\\
5. 开放性问题或建议性问题，比如：如何评价员工的绩效情况？\\
6. 涉及决策指导\\
7. 需要解释因果关系\\
8. 需要实时或动态数据\\
9. 需要深度分析或复杂模型

判断时应该考虑：问题是否有明确的答案，以及是否能仅通过已有的
历史数据得出这个答案。如果问题模糊或需要额外信息和分析，则应
判定为不能回答。

\#\#\# 以json格式输出：

\begin{verbatim}
{
    "label": boolean,   
    "tables": [          
        {
            "table_name": string,
            "fields": [string]  
        }
    ],
    "reason": string 
}
\end{verbatim}
\end{tcolorbox}
\end{CJK*}

\noindent\textbf{English version.}

\begin{tcolorbox}[breakable, colback=gray!10, colframe=blue!40, title=Prompt Template (English)]
It is 2025. You are a machine that writes SQL. You do not have 
data-analysis capability. Do not provide opinions for open-ended 
questions.

Your task is to judge, with the strictest standard, whether the 
user's question is a SQL-query request, whether it has a 
well-defined computational logic, and whether a definite answer 
can be obtained by generating and executing SQL within the given 
database scope.

\textcolor{magenta}{\{database\_meta\}}

\textbf{\#\#\# User Query:}

\textcolor{magenta}{\{user\_query\}}

\textbf{\#\#\# Instructions: Carefully analyze the user query and the database 
information, and answer following the examples below.}

Answerable questions usually have the following characteristics:

1. Query a specific historical data point, e.g., price or volume at 
   a certain time.\\
2. Perform simple statistical computations on historical data, e.g., 
   average, sum, max/min.\\
3. Retrieve records that satisfy specific conditions, e.g., values 
   above a threshold.\\
4. Query basic attributes or identifiers of an entity, e.g., code 
   or name.\\
5. Perform the above within a specified time range.

Unanswerable questions usually have the following characteristics:

1. No concrete and direct computational logic (e.g., How to analyze 
   employees' promotion paths?).\\
2. Future prediction or trend judgment (e.g., Will the employee 
   resign in the future?).\\
3. Subjective analysis or evaluation (e.g., How is the employee's 
   work capability?).\\
4. Require information beyond the database.\\
5. Open-ended or advice-seeking questions (e.g., How to evaluate 
   the employee's performance?).\\
6. Decision-making guidance.\\
7. Require causal explanation.\\
8. Require real-time or dynamic data.\\
9. Require deep analysis or complex models.

When judging, consider whether the question has a clear answer and 
whether the answer can be derived solely from the existing 
historical data. If the question is vague or requires additional 
information and analysis, it should be judged as unanswerable.

\textbf{\#\#\# Output in JSON format:}

\begin{verbatim}
{
    "label": boolean,      
    "tables": [            
        {
            "table_name": string,
            "fields": [string] 
        }
    ],
    "reason": string
}
\end{verbatim}
\end{tcolorbox}

\section{Additional Ablation Studies}
\label{sec:appendix_ablation}

\subsection{Robustness to Training Strategy}

\begin{table}[ht]
\centering
\small
\begin{tabular}{lcc}
\toprule
\textbf{Loss Function} & \textbf{F1 (\%)} & \textbf{AUC (\%)} \\
\midrule
\rowcolor{gray!25}
Label Smoothing ($\epsilon=0.1$) & \textbf{87.1} & \textbf{88.7} \\
Label Smoothing ($\epsilon=0.05$) & 85.1 & 88.4 \\
Focal Loss ($\gamma=2$) & 85.3 & 88.3 \\
Focal Loss ($\gamma=1$) & 85.2 & 86.9 \\
BCE Loss & 84.8 & 87.8 \\
\bottomrule
\end{tabular}
\caption{Loss function ablation on Qwen-3-8B / TriageSQL (4-layer TRGE, layer $-16$, dropout 0.2). Label smoothing with $\epsilon{=}0.1$ yields the best F1 (87.1\%) and AUC (88.7\%), outperforming standard BCE by +2.3\% F1, likely by mitigating overconfident predictions near the decision boundary.}
\label{tab:loss_ablation}
\vspace{-1em}
\end{table}

\begin{table}[ht]
\centering
\small
\begin{tabular}{lcc}
\toprule
\textbf{Dropout Rate} & \textbf{F1 (\%)} & \textbf{AUC (\%)} \\
\midrule
Dropout = 0.0 & 85.4 & 88.0 \\
Dropout = 0.1 & 86.5 & 88.4 \\
\rowcolor{gray!25}
Dropout = 0.2 (default) & \textbf{87.1} & \textbf{88.7} \\
Dropout = 0.3 & 85.5 & 88.0 \\
\bottomrule
\end{tabular}
\caption{Dropout rate ablation on Qwen-3-8B / TriageSQL (4-layer TRGE, layer $-16$, label smoothing $\epsilon{=}0.1$). Performance is stable across rates 0.1--0.3, peaking at 0.2 (87.1\% F1). The narrow variance ($\pm$1.6\% F1) indicates the probe learns distributed rather than brittle features.}
\label{tab:dropout_ablation}
\vspace{-1em}
\end{table}

\section{Baseline Experimental Settings}
\label{sec:baseline_settings}

We evaluate all baselines on Qwen-3-8B and Llama-3.1-8B backbones using bfloat16 precision, with a maximum sequence length of 2048 tokens. 

\subsection{Implementation Details}

\begin{itemize}
    \setlength\itemsep{0em}
    \item \textbf{Self-Evaluation ($P(\text{True})$):} We convert the question and schema into a declarative statement and prompt the model to judge its correctness using a True/False format. The logit of the ``True'' token is used as the answerability score. We use 4-shot prompting (2 answerable, 2 unanswerable) for calibration.
    \item \textbf{Semantic Entropy:} We sample $K=10$ outputs ($T=0.7$) and cluster them into semantic equivalence classes using an NLI model. For Text-to-SQL, refusal responses are treated as equivalent, while SQL queries are compared based on syntactic agreement. The entropy over equivalence classes serves as the uncertainty score.
    \item \textbf{CCS (Contrast-Consistent Search):} We construct contrast pairs (e.g., ``Is correct? Yes/No'') and extract hidden states from the last token of the final layer. An unsupervised linear probe is trained to satisfy consistency and informative constraints.
    \item \textbf{SAPLMA:} We extract the hidden states of the last token across all layers. A multi-layer perceptron (MLP) classifier ($4096 \to 256 \to 128 \to 64 \to 1$) is trained on a specific layer, selected via grid search on the validation set.
    \item \textbf{HaloScope:} We leverage an exemplar set to construct a hallucination-related subspace using Singular Value Decomposition (SVD). Answerability is predicted by training a classifier on the projection scores within this subspace.
    \item \textbf{TSV (Truthfulness Separator Vector):} We learn a separator direction in the latent space through consistency constraints. The answerability score is derived from prototype similarity after activation intervention.
    \item \textbf{Eigenscore:} We analyze the spectral statistics of hidden states across multiple samples ($N \ge 5, T=0.7$). The eigenvalue distribution of the activation covariance matrix is used to estimate epistemic uncertainty.
    \item \textbf{API-based Prompting:} We use DeepSeek-Chat with zero-shot instructions to directly judge query answerability. Refusal signals are extracted from the text output.
\end{itemize}

\providecommand{\textscmath}[1]{\ifmmode\text{\textsc{#1}}\else\textsc{#1}\fi}

\section{Algorithm Pseudocode}
\label{appendix:pseudocode}
\textit{Note: Algorithms \ref{alg:refusal_framework} and \ref{alg:refusal_training} describe optional extensions for combined hallucination detection and multi-domain adaptation, which are not part of the main results reported in Table 1.}

\subsection{Hidden State Extraction Framework}

\begin{algorithm}[!ht]
\caption{Hidden State Extraction for Answerability/Refusal Detection}
\label{alg:hidden_extraction}
\begin{algorithmic}[1]
\REQUIRE Dataset $\mathcal{D} = \{(q_i, c_i, y_i)\}_{i=1}^N$, LLM $\mathcal{M}$, Layer index $\ell$
\ENSURE Hidden state dataset $\mathcal{H} = \{(\mathbf{H}_i, y_i)\}_{i=1}^N$

\STATE Initialize empty dataset $\mathcal{H} \leftarrow \emptyset$
\FOR{each $(q_i, c_i, y_i) \in \mathcal{D}$}
    \STATE $\mathbf{x}_i \leftarrow \textscmath{Tokenize}(\textscmath{Prompt}(q_i, c_i))$
    \STATE $\{\mathbf{H}^{(l)}\}_{l=0}^{L} \leftarrow \mathcal{M}(\mathbf{x}_i)$ \COMMENT{Forward pass with hidden states}
    \STATE $\mathbf{H}_i \leftarrow \mathbf{H}^{(\ell)} \in \mathbb{R}^{T \times d}$ \COMMENT{Extract layer $\ell$}
    \IF{$\textscmath{HasNaN}(\mathbf{H}_i)$}
        \STATE $\mathbf{H}_i \leftarrow \textscmath{NanToNum}(\mathbf{H}_i)$ \COMMENT{Numerical stability}
    \ENDIF
    \STATE $\mathcal{H} \leftarrow \mathcal{H} \cup \{(\mathbf{H}_i, y_i)\}$
\ENDFOR
\RETURN $\mathcal{H}$
\end{algorithmic}
\end{algorithm}


\subsection{TRGE Probe Architecture}

\begin{algorithm}[!ht]
\caption{TRGE Probe for Answerability/Refusal Detection}
\label{alg:transformer_probe}
\begin{algorithmic}[1]
\REQUIRE Hidden state $\mathbf{H} \in \mathbb{R}^{T \times d_{in}}$, Padding mask $\mathbf{M} \in \{0,1\}^T$
\ENSURE Answerability probability $\hat{p} \in [0,1]$

\STATE $\mathbf{H} \leftarrow \textscmath{LayerNorm}(\mathbf{H})$ \COMMENT{Input Projection}
\STATE $\mathbf{Z} \leftarrow \textscmath{GELU}(\textscmath{LayerNorm}(\mathbf{H} \mathbf{W}_{proj} + \mathbf{b}_{proj}))$ \COMMENT{$\mathbf{Z} \in \mathbb{R}^{T \times d}$}
\STATE $\mathbf{Z} \leftarrow \mathbf{Z} + \textscmath{LearnablePE}(T)$ \COMMENT{Positional Encoding}

\FOR{$n = 1$ to $N$}
    \STATE $\mathbf{U} \leftarrow \mathbf{Z} + \textscmath{MultiHeadAttn}(\textscmath{LN}(\mathbf{Z}), \textscmath{LN}(\mathbf{Z}), \textscmath{LN}(\mathbf{Z}), \mathbf{M})$ \COMMENT{Residual 1: Attention}
    \STATE $\mathbf{V} \leftarrow \mathbf{U} + \textscmath{GELU}(\textscmath{LN}(\mathbf{U}) \mathbf{W}_1) \mathbf{W}_2$ \COMMENT{Residual 2: FFN}
    \STATE $\mathbf{Z} \leftarrow \mathbf{V} + \textscmath{SwiGLU}(\textscmath{LN}(\mathbf{V}))$ \COMMENT{Residual 3: Gating}
\ENDFOR

\STATE $L_{valid} \leftarrow \sum_{t=1}^{T} (1 - M_t)$ \COMMENT{Mean Pooling}
\STATE $\mathbf{z}_{pool} \leftarrow \frac{1}{L_{valid}} \sum_{t=1}^{T} (1 - M_t)\, \mathbf{Z}_t$
\STATE $\hat{p} \leftarrow \sigma(\textscmath{GELU}(\mathbf{z}_{pool} \mathbf{W}_{c1}) \mathbf{W}_{c2})$ \COMMENT{Classification Head}

\RETURN $\hat{p}$
\end{algorithmic}
\end{algorithm}


\subsection{Training Procedure}

\begin{algorithm}[!ht]
\caption{Distributed Training with F1-based Selection}
\label{alg:training}
\begin{algorithmic}[1]
\REQUIRE Training set $\mathcal{H}_{train}$, Validation set $\mathcal{H}_{val}$, Epochs $E$
\ENSURE Trained parameters $\theta^*$

\STATE Initialize $\theta, F_1^{best} \leftarrow 0, \theta^* \leftarrow \theta$
\FOR{$e = 1$ to $E$}
    \FOR{each mini-batch $\{(\mathbf{H}_i, y_i)\}_{i=1}^B \in \mathcal{H}_{train}$}
        \STATE $\mathcal{L} \leftarrow -\frac{1}{B} \sum_{i=1}^{B} [ y_i \log f_\theta(\mathbf{H}_i) + (1-y_i) \log(1-f_\theta(\mathbf{H}_i)) ]$
        \STATE $\theta \leftarrow \theta - \eta \nabla_\theta \mathcal{L}$
    \ENDFOR
    \STATE $F_1 \leftarrow \textscmath{F1Score}(\textscmath{Evaluate}(f_\theta, \mathcal{H}_{val}))$
    \IF{$F_1 > F_1^{best}$}
        \STATE $F_1^{best} \leftarrow F_1, \theta^* \leftarrow \theta$
    \ENDIF
\ENDFOR
\RETURN $\theta^*$
\end{algorithmic}
\end{algorithm}


\subsection{Refusal-Aware Hallucination Detection}

\begin{algorithm}[!ht]
\caption{Refusal-Aware Query Classification}
\label{alg:refusal_framework}
\begin{algorithmic}[1]
\REQUIRE Query $q$, Context $c$, LLM $\mathcal{M}$, Probe $f_\theta$, Thresholds $\tau_r, \tau_h$
\ENSURE Decision $\in \{\textscmath{Answer}, \textscmath{Refuse}, \textscmath{Hallucination}\}$

\STATE $p_{ans} \leftarrow f_\theta^{ref}(\mathcal{M}(\textscmath{Tok}(\textscmath{RefPrompt}(q, c)))^{(\ell)})$
\IF{$p_{ans} < \tau_r$} \RETURN \textscmath{Refuse} \ENDIF

\STATE $r \leftarrow \mathcal{M}.\textscmath{Generate}(\textscmath{Tok}(\textscmath{AnsPrompt}(q, c)))$
\STATE $p_{hal} \leftarrow f_\theta^{hal}(\mathcal{M}(\textscmath{Tok}(\textscmath{AnsPrompt}(q, c)) \oplus r)^{(\ell)})$

\IF{$p_{hal} > \tau_h$} \RETURN \textscmath{Hallucination} \ELSE \RETURN \textscmath{Answer} \ENDIF
\end{algorithmic}
\end{algorithm}


\subsection{Multi-Domain Refusal Training}

\begin{algorithm}[!ht]
\caption{Multi-Domain Refusal Training}
\label{alg:refusal_training}
\begin{algorithmic}[1]
\REQUIRE Domain datasets $\{\mathcal{D}_k\}_{k=1}^{K}$, LLM $\mathcal{M}$, Layer $\ell$
\ENSURE Domain-specific probes $\{f_{\theta_k}\}_{k=1}^{K}$

\FOR{each domain $k \in \{1, \ldots, K\}$}
    \STATE $\mathcal{H}_k \leftarrow \textscmath{ExtractHiddenStates}(\mathcal{D}_k, \mathcal{M}, \ell)$
    \STATE $\theta_k \leftarrow \textscmath{TrainProbe}(\mathcal{H}_k)$ \COMMENT{Algorithm~\ref{alg:training}}
    \STATE $F_1^{(k)} \leftarrow \textscmath{Evaluate}(f_{\theta_k}, \mathcal{H}_k^{test})$
\ENDFOR
\RETURN $\{f_{\theta_k}\}_{k=1}^{K}$
\end{algorithmic}
\end{algorithm}


\subsection{Model Configuration}

\begin{table}[!ht]
\centering
\caption{TRGE Probe Hyperparameters}
\label{tab:hyperparameters}
\begin{tabular}{lc}
\toprule
\textbf{Hyperparameter} & \textbf{Value} \\
\midrule
Input dimension ($d_{in}$) & 4096 \\
Model dimension ($d$) & 512 \\
Number of attention heads & 8 \\
Number of TRGE layers ($N$) & 4 \\
SwiGLU intermediate dimension & 2048 \\
Feed-forward dimension & 4096 \\
Dropout rate & 0.2 \\
Maximum sequence length & 8192 \\
Epochs & 20 \\
Optimizer & AdamW \\
\bottomrule
\end{tabular}
\end{table}

\section{Extended Case Study}
\label{sec:appendix_case_study}

\begin{figure}[ht]
    \centering
    \includegraphics[width=0.6\columnwidth]{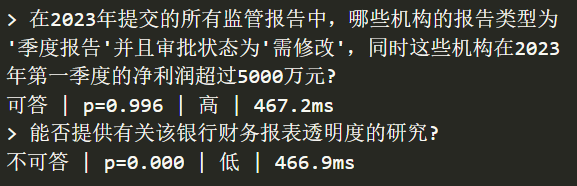}
    \caption{Running screenshot of \textsc{LatentRefusal} in a production Chinese financial QA deployment (NVIDIA RTX 4090, Qwen-3-8B backbone; hyperparameters optimized for production). \textbf{Top:} a multi-constraint regulatory query is correctly identified as answerable ($p{=}0.996$). \textbf{Bottom:} a subjective, out-of-scope request about ``financial statement transparency'' is correctly refused ($p{=}0.000$). End-to-end inference latency is stable at $\approx$467ms regardless of the gating decision. English translations of the Chinese queries are provided below.}
    \label{fig:case_study}
\end{figure}

We provide an extended case study to demonstrate \textsc{LatentRefusal}'s behavior in a realistic financial analysis scenario. Figure~\ref{fig:case_study} illustrates the model's response to two distinct types of user queries.

The two queries shown in Figure~\ref{fig:case_study} are translated as follows:
\begin{itemize}[leftmargin=*]
    \setlength\itemsep{0em}
    \item \textbf{Query 1 (Answerable, $p{=}0.996$):} ``Among all regulatory reports submitted in 2023, which institutions have a report type of `Quarterly Report' and an approval status of `Requires Revision', whilst also having a net profit exceeding 50 million yuan in Q1 2023?''
    \item \textbf{Query 2 (Unanswerable, $p{=}0.000$):} ``Can you provide information regarding the transparency of the bank's financial statements?''
\end{itemize}

In the first example (top), the user asks a highly specific question with multiple filtering conditions (``quarterly reports'', ``status needs revision'', ``net profit > 50 million''). Despite the syntactic complexity, \textsc{LatentRefusal} detects strong grounding between the query constraints and the database schema, assigning a high answerability probability ($p=0.996$). This demonstrates that the probe is not easily confused by query length or logical depth.

In the second example (bottom), the user asks for ``research on financial statement transparency'' regarding a specific bank. This is a subjective, open-ended request that cannot be resolved by a structured SQL query against the bank's transactional or reporting database. Baseline models often hallucinate SQL queries that retrieve loosely related text fields (e.g., \texttt{remarks} columns). In contrast, \textsc{LatentRefusal} identifies the lack of schema alignment for the abstract concept of ``research'' and correctly predicts the query as unanswerable ($p=0.000$), effectively serving as a safety gate.

Notably, the inference latency remains consistent ($\approx 467$ms) regardless of the decision, confirming the efficiency of the single-pass architecture.

\end{document}